\documentclass{article}

\usepackage{arxiv}

\usepackage[utf8]{inputenc}
\usepackage[T1]{fontenc}
\usepackage{hyperref}
\usepackage{url}
\usepackage{graphicx}
\usepackage{amsmath}
\usepackage{microtype}
\usepackage{csquotes}
\usepackage[english]{babel}
\usepackage{natbib}
\usepackage{doi}

\hypersetup{
  colorlinks = true,
  linkcolor  = blue,
  citecolor  = blue,
  urlcolor   = blue
}

\renewcommand{\headeright}{Post-print $\cdot$ \textit{Synthese} 205:137 (2025)}
\renewcommand{\undertitle}{Post-print $\cdot$ \textit{Synthese} (2025) \textbf{205}:137 $\cdot$ \href{https://doi.org/10.1007/s11229-025-04961-4}{doi:10.1007/s11229-025-04961-4}}
\renewcommand{\shorttitle}{Generative Midtended Cognition and Artificial Intelligence}

\title{Generative Midtended Cognition and Artificial Intelligence\\[4pt]
{\large\textit{Thinging with Thinging Things}}%
\thanks{This is the authors' post-print (accepted manuscript) of an article
published in \textit{Synthese} (2025) \textbf{205}:137.
The final authenticated published version is available open access at
\url{https://doi.org/10.1007/s11229-025-04961-4}.
\copyright~The Author(s) 2025. Open access under a Creative Commons
Attribution (CC~BY) licence.
Received: 30 July 2024; Accepted: 9 February 2025.}}

\author{%
  Xabier E.~Barandiaran\textsuperscript{1} \quad \& \quad
  Marta P\'erez-Verdugo\textsuperscript{1,*}\\[4pt]
  \textsuperscript{1}IAS-Research Centre for Life, Mind and Society,\\
  Dept.\ Philosophy, UPV/EHU, University of the Basque Country,\\
  Donostia, Gipuzkoa (Spain)\\[4pt]
  \href{http://xabier.barandiaran.net}{\texttt{xabier.barandiaran.net}}
  \quad \href{mailto:xabier.barandiaran@ehu.eus}{\texttt{xabier.barandiaran@ehu.eus}}\\
  ORCID: \href{https://orcid.org/0000-0002-4763-6845}{0000-0002-4763-6845}\\[2pt]
  \href{mailto:marta.perezv@ehu.eus}{\texttt{marta.perezv@ehu.eus}}\\
  ORCID: \href{https://orcid.org/0000-0003-2515-0462}{0000-0003-2515-0462}\\[4pt]
  \textsuperscript{*}Corresponding author
}

\date{}

\begin{document}

\maketitle

\begin{abstract}
This paper introduces the concept of ``generative midtended cognition'',
that explores the integration of generative AI technologies with human
cognitive processes. The term ``generative'' reflects AI's ability to
iteratively produce structured outputs, while ``midtended'' captures the
potential hybrid (human-AI) nature of the process. It stands between
traditional conceptions of \textit{in}tended creation, understood as
steered or directed from with\textit{in}, and \textit{ex}tended
processes that bring exo-biological processes into the creative process.
We examine the working of current generative technologies (based on
multimodal transformer architectures typical of large language models
like ChatGPT), to explain how they can transform human cognitive agency
beyond what the conceptual resources of standard theories of extended
cognition can capture. We suggest that the type of cognitive activity
typical of the coupling between a human and generative technologies is
closer (but not equivalent) to social cognition than to classical
extended cognitive paradigms. Yet, it deserves a specific treatment. We
provide an explicit definition of \textit{generative midtended
cognition} in which we treat interventions by AI systems as constitutive
of the agent's intentional creative processes. Furthermore, we
distinguish two dimensions of generative hybrid creativity: 1.
\textbf{Width}: captures the sensitivity of the context of the
generative process (from the single letter to the whole historical and
surrounding data); 2. \textbf{Depth}: captures the granularity of
iteration loops involved in the process. Generative midtended cognition
stands in the middle depth between \textit{conversational} forms of
cognition in which complete utterances or creative units are exchanged,
and micro-cognitive (e.g.\ neural) subpersonal processes. Finally, the
paper discusses the potential risks and benefits of widespread generative
AI adoption, including the challenges of authenticity, generative power
asymmetry, and creative boost or atrophy.
\end{abstract}


\section*{Acknowledgements}

XEB and MPV acknowledge IAS-Research group funding IT1668-22 from Basque
Government, grants PID2019-104576GB-I00 for project Outonomy, and
PID2023-147251NB-I00 for project Outagencies funded by
MCIU/AEI/10.13039/501100011033. XEB acknowledges project MANTIS, Grant
01621-01493911 MANTIS also funded by MICIU/AEI/10.13039/501100011033 and
by the European Union NextGenerationEU/PRTR. MPV is developing her PhD
thanks to the grant PRE2020-096494 of the Spanish National Research
Agency (AEI) and cofunded by the European Social Fund.

\section{Introduction}
\label{sec:introduction}

\begin{quote}
\textit{``I really do think with my pen, because my head often knows
nothing about what my hand is writing''.}\\
Wittgenstein (1980, p.~106)
\end{quote}

You are surrounded by colleagues in a conference. You are about to
explain your opinion about the French philosopher Gilles Deleuze: ``He
is very inspiring, but his writing is too \ldots'', ---you can't quite
find the right word, ``\ldots\ abstruse'' says your colleague, ``yeah,
his writing is too abstruse (thanks)'' you continue. That is the word
you needed, the one you wanted but could not find. You do in fact hold
the opinion that Deleuze is abstruse, you simply could not generate the
sentence fluently, and you completed it by accepting the offered
suggestion. You made it yours. At the current pace of evolution of
generative technologies, it is not unreasonable to suggest a scenario
in which similar conversations are increasingly generated (suggested) by
AI (instead of your colleagues). How would this be possible? What kind
of cognitive process would this be? How are this and parallel scenarios
different to any technologically or socially extended cognitive
processes we experienced before? How should we characterize them?

Generative technologies, and Large Language Models in particular
(systems like ChatGPT, Gemini, Llama, Mixtral, Claude, etc.), are
deeply transforming agency. In a recent article, \citet{BarandiAran2024arxiv}
introduce the concept of \textit{midtension} to characterize the type
of hybrid ``intentional'' agency that can result from a deep integration
between human and LLM interaction:

\begin{quote}
``Transformers are also bringing with them a much deeper meaning of
extended agency (with deeper dialectical connotations). There is a form
of extended agency that LLMs already offer that get more intentionally
intimate than any previous known form. In fact, this extensional
character is closer to the intentional character of the mind that
deserves a proper name: midtensional. (\ldots) The enormous complexity
and regulatory capacity of the brain-body system (compared to that of
the passive materiality of the tool and work environment) is now
challenged by an ongoing activity of language automata, which are
constantly reading us and writing (for) us. (\ldots) This brings the
power of transformer-human interaction closer to a proper cyborg agency,
beyond any experience of instrumental, social or intersubjective agency
we might have ever encountered before.''
\citep[pp.~29--30]{BarandiAran2024arxiv}
\end{quote}

In this paper we expand, deepen and generalise over this basic intuition
beyond text-based LLMs to generative technologies. We develop the notion
of ``generative midtended cognition'' as a new type of so-called
``extended cognition''---that is, processes that are characterised as
cognitive and are constituted by factors external to the cognizer's
brain-body. The meaning will unfold along the paper, but we shall
advance that we have chosen the term ``generative'' to mean what, in
different circumstances, might have been called ``creative'', yet devoid
of the strong connotations of the term. The term generative has, of
course, also been chosen to name the generative AI technologies that
have emerged recently \citep{Akhtar2024,Jebara2012,Murugesan2023}.
Altogether we want to stress processes that produce---generate---structured
material outcomes: text, drawings, sound, voice, shapes, etc.
Midtended or midtension is a neologism that wants to capture a space
situated between traditional conceptions of \textit{in}tention or
\textit{in}tended creation, that is, generated from within, and
\textit{ex}tended, processes that bring material exo-biological
processes into the creative process.

In the next section, we introduce so-called ``generative AI
technologies'' and their internal workings. Then, we argue that existing
theories of cognition that have incorporated external or environmental
components into cognitive processing fall short of adequately capturing
the new forms of cognition and agency that generative AI makes possible.
Section~\ref{sec:midtension} introduces the concept of generative
midtension with the examples of drawing and writing. We then articulate
the relationship between the concepts of intention and extension and
characterise the singularity of midtended cognition. We provide an
explicit definition of midtended cognition and distinguish two
fundamental dimensions along which generative midtended cognition can be
demarcated. Section~\ref{sec:future} introduces some future scenarios
that are relevant to deeper senses of generative midtended cognition,
we evaluate some potential benefits and risks of authenticity,
generative power asymmetry, and creative atrophy and alienation.
Finally, Section~\ref{sec:conclusion} recapitulates on the main ideas
of the paper and offers some concluding remarks.

\section{Extending Extended Cognition: Generative AI's New Challenges}
\label{sec:extending}

\subsection{The Era of Generative AI}
\label{subsec:era}

Generative Artificial Intelligence has reached maturity and widespread
adoption after transformer architectures first managed to deliver highly
proficient text generation \citep{Vaswani2017}. Today it is possible for
anyone to access AI services that can generate text, image, audio and
video, and whose quality is often indistinguishable from that created by
humans. To characterise these systems, it is important to understand in
some detail their internal workings \citep[see][for a more detailed
philosophical analysis]{BarandiAran2024arxiv}.

Generative systems take as input a string or matrix of data (a text
document, audio file, image, or combinations of them) and break them
down into tokens or basic processing units. The tokenized input is then
transposed to an intermediate structured ``representation''
(\textit{embedding} in the case of language or \textit{latent space} in
the case of image diffusion models) that takes the shape of a matrix in
a high dimensional space. Attention mechanisms make it possible for
points of this high dimensional vector to be related to each other in
multiply structured manners (grammatical relationships between subjects
and verbs in a sentence, instruction steps or a receipt, continuity of
shape in a picture). Neural networks take over now. They process the
matrix by transforming it further through massive non-linear distributed
computations that are said to express ``knowledge'' or to constitute a
model of the training data (see below). Attentional and neural
processing blocks (96 in the case of GPT-3, see \citealt{Brown2020})
are chained in this processing line. The resulting matrix is one that
assigns a weighted probability to each possible token. An algorithm then
picks up one among the highest probability output tokens. The whole
process is mediated by a processing architecture composed of billions of
parameters.

Importantly, the generation of a complete output (paragraph, full text,
complete image, etc.) is not something created in ``one shot'' but
recursively, token by token, pixel by pixel, by feeding the system with
its own output (where the human agent can intervene, by correcting,
suggesting, or adding new tokens to the input). Whereas LLMs'
\textit{autoregressive} dynamics rely on reintroducing the output token
into the new input (adding the new token to the previous input),
generative diffusion models work on a kind of internalised noisy sketch
that gets iteratively completed by a similar process of modifying pixels
while processing previous image and text (or even sound) input (like
image instruction).

In its raw initial state, a generative AI system is almost a blank
slate, waiting to be shaped by the training data to determine embedding,
attentional and neural parameters. The basic training procedure is to
provide the system with an input, to let the system process it, and then
to compare the result with the next token (word) of the training data.
If the training data includes Clark and Chalmers' ``The extended mind''
paper, the training process will include delivering as an input pieces
of the text, like the beginning of the paper ``Where does the mind stop
and the rest of the world [\ldots]'', and letting the system ``guess''
the next word. During training, the result will range from absolutely
random at the beginning (e.g.\ ``Paris'') to an approximate bad guess
in earlier stages (e.g.\ ``rest''), to a reasonable match (e.g.\
``start''). The output is then compared to the next word on the original
text: ``begin''. The ``distance'' between the generated and the correct
output is used to train the system backwards, adjusting all the
parameters to reduce the error. So ``Paris'' will generate a big error,
but ``start'' a small one.\footnote{The real procedure is more
complicated (e.g.\ all output's error is 1 except for the ``correct''
term) but the net effect of the error propagation will be smaller for
terms that are closer in the embedding space (e.g.\ ``start'' and
``begin'').}

In this sense, the system is trained to predict the most likely
follow-up sequence of a given input (often called prompt). Sixtillion
operations of error backpropagation and parameter fitting have been
shown to deliver very efficient systems able to reasonably approximate
training data. But, most importantly, able to generate reasonable
completions to previously unseen data. To the input ``Deleuze is very
inspiring, but his writing is too'' (never seen in the training phase),
ChatGPT responds with ``dense and abstract for some readers.''%
\footnote{\url{https://chatgpt.com/share/8a56c0cb-95fc-48ed-aa01-b8d2ec6adac4}}
Furthermore, foundation models (as they are called after the first
intensive training phase) can be fine-tuned for specific tasks or to
adapt their generic abilities to specific content, styles, or different
types of customizations. It is possible to tune ChatGPT (and other
LLMs) with personal conversation records, emails, messaging conversation
and papers to fit my personality and complete the above sentence with
``abstruse''. In addition, the input to a generative system can include
contextual information (like the set of papers you want to reference in
a paper you are about to start writing, or all the emails you exchanged
with a particular person, or a full genocide-victim database).%
\footnote{And here is where hallucinations, biases and reasoning errors
are still preventing a generalised rolling down into the generative
road.}

Being the fastest growing digital service in history \citep{Hu2023},
ChatGPT and similar generative technologies producing text (Gemini,
LLama, Mixtral, Claude, etc.), images (Dall-e, Midjourney, Stable
Diffusion), music (Udio, Aiva, WavTool) or video (Sora, Pika Labs, or
Runway) are part of our cognitive resources. Moreover, generative AI
services, increasingly involve the disposal of joint human-machine
creative digital spaces (Canvas on ChatGPT, NotebookML on Gemini,
Artefacts on Claude or NotebookLlama) and complementarily, digital
productive environments (from Gdocs and Gmail to the Microsoft or Adobe
suites) incorporate generative AI support for a variety of creative
tasks (from email writing to image editing). What are our theoretical
resources to accommodate these tools and environments to our best
understanding of how human cognition emerges out of recurrent
agent-environment interactions?

\subsection{Extended Cognition Theories and the Challenge of
Generative Environments}
\label{subsec:extended}

Many philosophical accounts, particularly during the 20th century, have
emphasised the role of the technical environment in human cognition and
experience. Notable examples include Haraway's cyborgs and situated
knowledge \citep{Haraway1991}, Stiegler's Heideggerian account of
technique \citep{Stiegler1998} and Hutchins' distributed cognition
approach \citep{Hutchins1995}. However, the extended mind hypothesis
\citep{Clark1998extended} marks the first significant attempt, within
philosophy of mind, to strongly argue on the triviality of retaining the
mind within the boundaries of the skull. This hypothesis, rooted in a
functionalist philosophy of mind, coined the parity principle as the
main support for its argument: if a process considered cognitive when
performed by the brain can also be instantiated including elements of
the environment, then those elements are cognitive to the same extent as
the brain.

Clark and Chalmers illustrate this with the example of Inga, a healthy
individual, and Otto, an Alzheimer's patient. Both individuals
successfully perform the same cognitive process (remembering the address
of their favourite museum), but in each case it is instantiated
differently---Inga using only her brain and Otto using his brain and a
notebook. Otto has an extended mind. To clarify this claim, Clark \&
Chalmers propose some criteria that environmental resources must meet to
qualify as cognitive extensions: trustworthiness, reliability and
accessibility. Known as the ``trust \& glue'' criteria, they seem to
follow the intuition that the less controversial cases of extension are
prostheses or bodily extensions; fixed and static environmental
resources that the agent always carries with herself.\footnote{We have
chosen to mix both in a single dimension for simplicity's sake. However,
we could be more precise and differentiate between two dimensions,
keeping the label ``Width'' for the dimension covering the system's
sensitivity to the particular context in which the generative process is
situated, and adding a second dimension labelled ``Length'' that covers
the system's sensibility to the specific agent's previous history of
creative processes and outputs, displaying a temporal character.}

After a first wave focused on defending the possibility of extended
minds, a second wave of theorists moved away from strict functionalism
and the stringent parity principle \citep{Sutton2010}. Instead, they
adopted complementarity-based accounts, where external resources
integrated in cognitive processes need not mirror brain functions
exactly, fostering different kinds of cognitive extension based on
varying dimensions of integration. \citet{Heersmink2015} offers a
taxonomy of these dimensions as presented in the literature, retaining
some of the early intuitions of durability, reliability and trust,
reformulating others such as phenomenological and informational
transparency, and including newly proposed dimensions such as
informational flow, individualization and transformation.\footnote{We
tentatively term the kind of integration we are targeting ``active''
integration, instead of the more passive kind of integration theorised
in extended cognition so far. We are aware, however, that this might be
misleading as, in a different context, extended cognition was precisely
characterised as a sort of ``active externalism'' against the ``passive
externalism'' of externalist semantics \citep{Hurley2010}. We have
chosen to maintain it like that for now, given that the ``active''
etiquette there is used in a sufficiently different comparison and
terrain.} This allows a more nuanced and fruitful description of
extended cognition.

But it is still noteworthy for our purposes that the most used examples
of extended cognitive processes in early theorising were memory
processes, where it is particularly relevant that the environmental
resources remain fixed and stable, thus making a certain passivity of
environmental elements the ideally characterised integrators of extended
cognitive processes (in open contrast with generative AI). In this
sense, Material Engagement Theory (MET) \citep{Malafouris2016} as a
follow-up of extended cognition theories that takes a more enactive and
dynamical approach to cognition, is clearer in the non-passive role that
it assigns to the environment. MET not only claims that cognition is
extended beyond the skull, but that it is dynamically co-shaped by
material interactions. Thinking is not previous to the encounter with
matter. Thinking is \textit{thinging}, doing things with matter.
\citet{Malafouris2019} uses the example of pottery (note the difference
in the choice of example, opting here for an uncontroversially creative
and active process). Contrary to the idea of the potter imposing a
predefined form onto passive matter (as if having a 3D model that is
then applied to the clay), the form starts to emerge in the process of
the potter manually interacting with the materiality of the clay. The
malleability of the clay in response to hand movements is what
constitutes the creative process. Whereas in traditional theorising of
mind the generative cognitive locus was ``in the head'' of the human
agent, here the generative locus of cognition is situated in the meeting
with the environment, driven both by material constraints and human
creative agency.

Extended cognition theories, as part of the 4E paradigm in cognitive
science, have proved to be rich frameworks to study contemporary digital
technology \citep{Smart2017}. AI-based technologies, however, pose
particular challenges, and ethical debates have questioned the
suitability of adopting extended cognition frameworks for these
contemporary technologies \citep{Cassinadri2022,Farina2022}. The
challenge stems from the intuition that extended cognitive processes
should retain a certain locus of control or authorship within the human
agent, so we can still consider it \textit{her} cognitive process. We
could make use of trust \& glue criteria and dimensions of integration
to explore these issues of control, but they turn out to be insufficient
in many cases, since they operate under the assumption that the
environment does not have any sort of generative power. Under this
supposition, it makes sense to think, for example, that the more
transparently we use an artefact, the more integrated it is in a
cognitive process: more transparency would mark that the artefact is
more directly \textit{responding} to the users' intentional agency (and
only to it), since the human agent is the only locus of generative,
purposeful interventions in the interaction.

But the intuition that more transparency equals more integration (in
this sense of clear authorship of behaviour) starts to break down when
we consider AI-powered technologies \citep{Clowes2020,Wheeler2019}.
\citet{PerezVerdugo2022} analyses this problem with the example of a
smart racket that adjusts your movements to optimise your play. Here, we
can have a very strong phenomenological feel of transparency, since we
can enter an uninterrupted flow of action. Nevertheless, the artifact is
not directly responding to our intentional agency, but is actively
anticipating it and \textit{intervening} on the joint action at a scale
the agent is not fully aware of. The artifact constitutes its own locus
of generative, purpose-\textit{structured} interventions in the
cognitive process that involves both agent and artifact. The generative
locus of cognition is now no longer only on the side of the agent in
its meeting with materiality. There is now a second locus in the
environment that can generatively alter the materiality of the
encounter.

The important novelty of Generative AI is that, on the pottery analogy
of material embodied cognition, clay is starting to shape itself by
anticipating our movements. And it does so based on the huge amounts of
data coming from: a)~the history of pottery and on all contemporary
catalogues, b)~on the current shape being moulded, c)~on our prior
declared intentions, d)~on the particular work context and, e)~on our
personal-style expressed on all our previous production. And we can
engage with it in ways that range from the AI inter-\textit{actively}
grasping and directing our hands, to simply letting us rest at a
distance to see the clay take shape. Thinking (cognition), might not be
anymore a process understandable as ``mere'' \textit{thinging}. Our
computational civilization has created \textit{thinging things}, that
is, things (programs) that recursively (autoregressively) transform
digitality (as a type of materiality) and generate other things
(structured digital objects, including other programs). We live, thus,
in an era where cognition can be extended in a very specific form, as a
very special form of thinking: \textit{thinging with thinging things}.

\section{Generative Midtension}
\label{sec:midtension}

\subsection{Intention and Authorship in Creative Action: Drawing and
Writing}
\label{subsec:drawing}

Imagine you are a professional designer. You received an email with a
client's request. You externalise your thoughts: ``This work requires an
80s style drawing with a contrasting combination of sharp and round
shapes''. Creating this work implies many recursive processes: from
looking at the blank piece of paper or screen and projecting imagined
figures, lines, shapes and compositions (maybe even using hands or other
objects to get a grasp of proportions), to sketching in pencil the
imagined future drawing, to inking it (making it more definitive) and to
finally colouring it. So, as \textit{you} start drawing, \textit{you}
get to imagine the completion of your traces on the white background as
a grey coloured (pencil) sketch. Sometimes \textit{you} augment the
focus of your anticipated sketch and keep imagining. Some other times,
\textit{you} directly intervene on it: \textit{you} either keep drawing
following the grey lines that \textit{you} drafted and continue
imagining them as \textit{you} draw them. But you also often choose to
diverge from the draft, and \textit{you} imagine how the picture
re-organizes accordingly. Sometimes you surprise yourself as \textit{you}
have sketched, on \textit{your} first try, exactly what you wanted. Some
other times, you take your distance, \textit{you} sketch again and keep
correcting till \textit{you} get it right.

We have all had similar extended creative experiences before. But with
generative AI, the role of imagination and of the material pencil sketch
(or even that of the ``inker''), can be played by a machine. By a
generative AI working in real-time \textit{within} our creative process.
Read the passage above again and substitute the ``you'' in italics by a
generative AI. Although still relatively fictional, this possibility is
increasingly foreseeable. It is partly a reality for computer
programmers making intensive use of code completion. And it is partly
available as autocompletion and other generative AI services embedded
within office environments or audiovisual creative and editing tools. We
explained above how these systems can be trained, tuned and contextually
fitted to a particular user, her past work and surrounding information.
The system could have read the client's request, taken the designer's
expressed intentions, and projected the style and features of her past
designs to this joint generative process; while actively adapting to the
designer's intentions in real-time.

In a similar vein, \citet{BarandiAran2024arxiv} provide the following
explanation of writing as a generative process:

\begin{quote}
``The process of writing (in paper or on the screen) is one that it is
often experienced as the very act of writing driving itself the
intentions of the writer: the interaction process of writing pulls
agency out of the head. It is the recurrent
hand-keyboard-PC-screen-vision-brain-body-hand loop that produces text.
Yet, we don't only write. We also supervise and edit recurrently. Thus,
at least two loops are involved here, one is more pulled by the direct
writer-text dynamics, the other by the more detached editorial
supervision that either continuously or intermittently follows the
former. At times, one finds the non consciously written text as right
and owned, as proper, and it is left untouched. Other times\ldots `That
is not what I meant exactly' ensues, `it needs a rewrite'. Both loops
are person anchored. The environment (pen and paper or keyboard and
screen) served as a support structure, a well integrated, creative
scaffold, providing the material basis of extended memory, recomposition,
and tinkering. But the writer was the extended agent, the organized
center of the scaffolded subject. This might start to change.''
\citep[p.~29]{BarandiAran2024arxiv}
\end{quote}

A variation of Hemingway's motto ``write drunk, edit sober'' is becoming
widespread in many areas of human text generation: ``let the LLM write,
edit yourself'' and very often the inverse ``write drunk, let the LLM
edit''. Both loops used to be person anchored, but now LLMs can
intervene on any or both of them and create, alter or shape (parts of)
the final creative product.

These are all examples of generative activities. We shall use the term
``generative'' as a more generic or neutral concept (regarding mindful
attribution) for what, under normal circumstances and performed by a
human, would be considered a \textit{creative} production process.
Unlike other forms of cognition (like decision-making, recalling,
navigating, imagining, judging, understanding)\footnote{Although it is
possible to do all these things generatively, they are not defined by
their generativity.} generative cognition is characterised by the
generation of a sequence of transformations on the environment to compose
a final product. Under most circumstances, generative processes are
creative, purposeful processes. Intentions take shape along the process
and the final result is authored. You identify yourself with the result.
When honest, a sense of authorship ensues, like the sense of agency that
accompanies intentional and voluntary action. But, with generative AI,
as we have seen, the environment becomes populated and shaped by digital
automata capable of producing (as Barandiaran and Almendros put it)
purpose-structured (although not fully purposeful) generative processes.

What happens when the generative process is hybrid? When machine
generated units (tokens in LLM technical vocabulary) are systematically
combined with human generated ones? When production and edition,
sketching and inking, are recurrently intertwined between machines and
humans? What happens when machine contributions are tuned to context and
author-style and knowledge? When the machine inputs augment, condition
and expand authorship?

\subsection{Intention, Extension and Midtension}
\label{subsec:midtension-def}

For classical approaches to human action, intentions are the result of
the right combination of beliefs and desires causing the action
\citep{Davidson1980}. This kind of event-causal account of
intentionality leaves little room for explaining generative midtended
action. $\varphi$ becomes an intentional action if $X$ (the agent) holds
the belief that doing $\varphi$ will satisfy desire $D$. Both $\varphi$
and $D$ are specified prior to the action and the desire $D$ and the
belief that ``$\varphi$ will bring $D$'' must be contained and causally
efficacious in the mind of $X$ in order to make $\varphi$ an intentional
action.

Contrary to the standard view, enactive and embodied contemporary
approaches to cognition \citep{Varela1991,Calvo2008,Clark1998,Gallagher2023}
stress the emergent nature of intentionality through sensory-motor
interactions between agents and their environments. There are various
families of such approaches. On the active inference or predictive coding
side \citep{Clark2013}, intentionality is understood as emergent from
recurrent predictive loops, where action is directed at reducing
uncertainty and updating the agent's future predictions of its
environment. As \citeauthor{BarandiAran2024arxiv} stress, from this
point of view, the novelty of generative technologies is that: ``we
might be encountering, for the first time, that the environment is
delivering to the brain-body the very predictions that the brain-body is
about to make about the effect of its own activity on the environment
(\ldots) In a textualized manner, this form of autocomplete is
equivalent to injecting predictive efferent signals into the body
movement.'' \citep[p.~30]{BarandiAran2024arxiv}.

Enactive, autonomy-centred sensorimotor approaches, on the other hand,
stress the importance of the agent-centred adaptive (i.e.\
norm-sensitive, purposeful) modulation of the sensorimotor coupling with
the environment \citep{Barandiaran2008,Barandiaran2017,Barandiaran2009,%
DiPaolo2017}. This approach assumes that the sensory-motor coupling is
modulated from within or centred on the agent's side. How to make sense
of the increasing active automaticity of the environment, producing
normatively-structured changes in the environment while coupled to the
activity of the agent? As the word-suggesting example in the opening of
the paper illustrates, generative cognition might be better understood
through the conceptual resources of social cognition \citep{Perez2021}
than those of extended cognition. In the enactive approach, this would
imply conceptualising generative cognition as a case of participatory
sense-making \citep{Cuffari2015,DeJaegher2007}. However, generative
technologies do not qualify as autonomous agents
\citep{BarandiAran2024arxiv,Gubelmann2024,Harvey2024}. Thus, it is
problematic to apply social cognition categories to interactions with and
between generative AI systems, if we make justice to their lack of
genuine autonomy, agency and understanding.

Technology, in turn, has also started to be analysed within enactive
theorising. In a recent paper, \citet{PerezVerdugo2023} define technical
behaviour as the active transformation and organisation of elements of
the body and environments ``with the intended effect of constraining or
regulating couplings with (or between) other aspects of the environment''
(p.~83), and technology is seen in this framework as the ``sedimented
effects of this technical behaviour at different scales within a systemic
context'' (p.~84). This does indeed underline the strong relation between
agency, productive practices \citep[see also][]{DiPaolo2023} and
technical interventions, and the enactive idea of coupling stresses the
co-constitutive and constraining role of the environment in cognitive
processes. However, this account still retains the idea that the
coupling with the (technological) environment is only
\textit{generatively} regulated by the agent---which is the only one
capable of technical behavior in the coupling. Enactivism, thus, has not
yet accommodated scenarios in which the environment is
\textit{generatively} norm-following or purpose-structured and tuned to
the autonomy of the human agents and the sociotechnical context in which
they operate.

Following \citet{BarandiAran2024arxiv}, we differentiate between the
purpose\textit{ful} (or intentional) behaviour characteristic of human
agents, and the merely purpose-\textit{structured} and
purpose-\textit{bounded} behaviour that generative AI produces. The
gigantic amounts of data that is used to train generative AI is the
result of human purposeful activity and the expression of human purpose.
Massive iterative parameter tuning techniques result in AI systems
becoming purpose-structured themselves \textit{and} able to generate
purpose-structured output. Moreover, the training process with
reinforcement learning (often with human feedback) channels the resulting
behaviour within the bounds of human expectations, desires, goals and
values, constraining its generative capacities within
purpose-bounded outputs.

To further clarify this point, we can make use of the distinction
between intrinsic and derived intentionality \citep{Searle1983}.
Generative AIs are not intrinsically intentional systems, but they
possess derived-intentionality, like other human creations (from simple
utterances to complex technologies). However, unlike previous signs and
technologies, this time the derived intentionality is also
\textit{generative of derived intentional products}. Imagine a set of
road signs in a city; they are certainly not intrinsically intentional
entities, not purposeful systems. The STOP signs do not understand
English, nor do they ``know'' that they are there to stop cars, or even
to signify anything. However, they are all purpose-structured and
purposefully-arranged, they operate as systems with derived (human
designed and repaired) intentions to avoid car crashes, to reduce car
speed and pollution, to direct the traffic towards the widest avenues,
and to reduce it in school areas. Now imagine that, taking a set of
initial prescriptions and based on an immense road traffic data, an AI
is capable of generating a similar kind of purpose-structured
traffic-sign distribution. We have here a second-order
derived-intentionality system (the generative AI), capable of generating
third order derived-intentional or purpose-structured products (texts,
images, etc.).

The framework we are developing in this paper allows for the integration
of both social-cognitive aspects and technological constitutive aspects
into a unified theory (both for enactivism and predictive coding
approaches). A proper characterization of the cognitive processes that
occur in interaction with generative AI is required for this task. As a
first step, we provide the following \textbf{definition of generative
midtension}:

\begin{quote}
Given a cognitive agent $X$, a generative system $Y$ (artificial or
otherwise) and cognitive product $Z$, midtension takes place when
generative interventions produced by $Y$ become constitutive of the
intentional generation of $Z$ by $X$, whereby $X$ keeps some sense of
agency or authorship over $Z$.
\end{quote}

\begin{figure}[ht]
  \centering
  \includegraphics[width=0.75\textwidth]{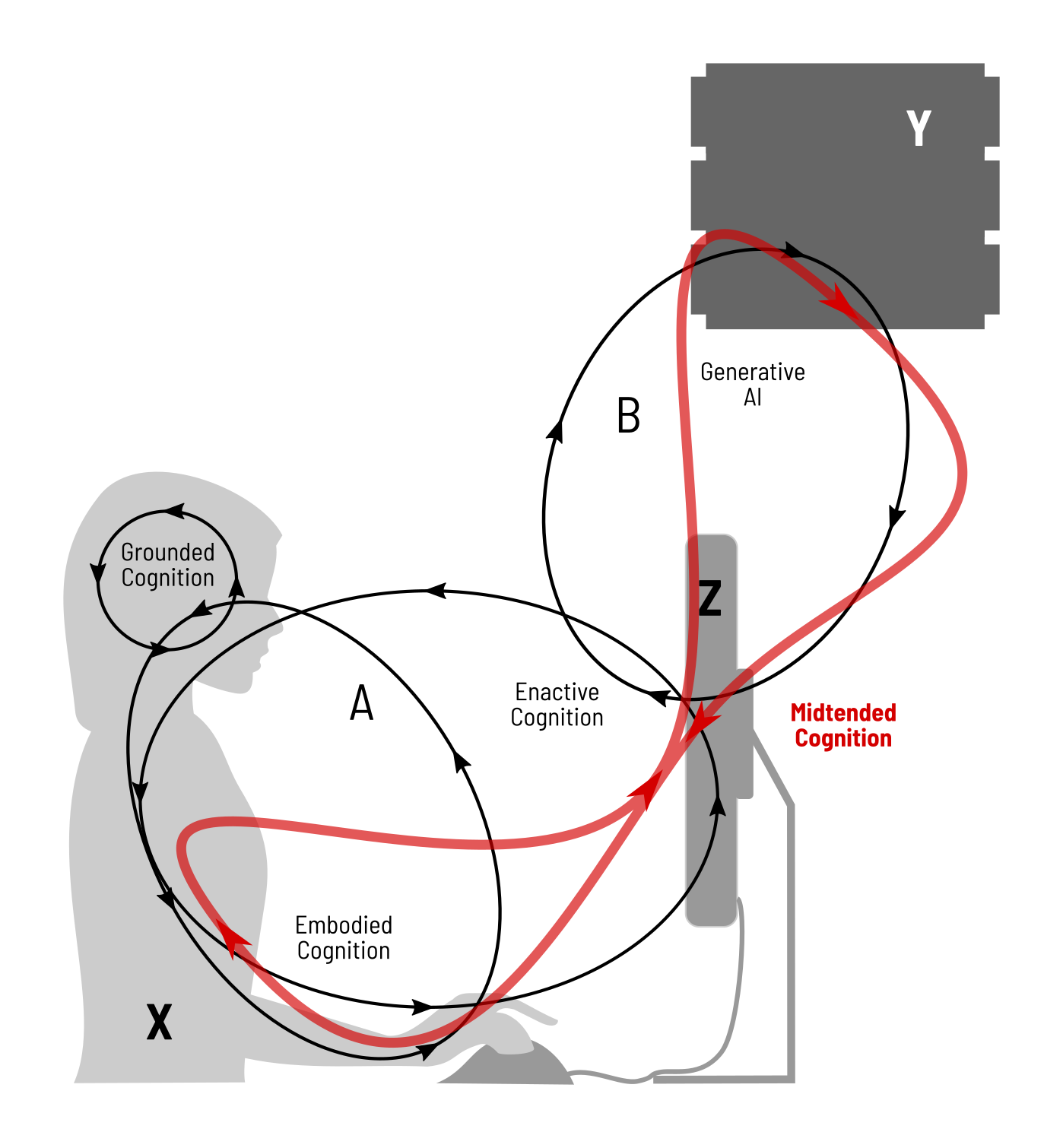}
  \caption{Different agent-environment loops generate different aspects
  of cognition. Midtended cognition involves the intersection between
  more standard agent anchored A-type loops and AI produced B-type
  loops. See text body for details. \textit{[Copyleft 2024
  \textcopyright\ Barandiaran and P\'erez-Verdugo, with CC-by-sa
  licence, the person's silhouette and the server image are derivative
  works of original SVG files of Vectorportal.com, with permission]}}
  \label{fig:loops}
\end{figure}

Figure~\ref{fig:loops} illustrates the definition and the various ways
in which grounded, embodied and extended cognitive processes interrelate
with each other and with generative processes on enacting midtended
cognition. Agent $X$ interacts with digital product $Z$, which is partly
generated by $Y$, an artificial generative technology.

Different agent-environment loops generate different aspects of
cognition: \textit{Grounded cognition} implies neurodynamic loops
involving multimodal pre-motor and sensory processing areas,
\textit{embodied cognition} implies body proprioceptive feedback and
neuromuscular synergies, \textit{enactive cognition} involves the
coupling with environmental elements, and \textit{generative AI} (not a
kind of cognition \textit{per se}) produces digital structures through
autoregressive loops. \textit{Generative midtended cognition} brings all
components together on the production of somewhat intentional, purposeful,
works.

Creative intentional loops through the agent (A loop) are nested with
the autoregressive loops of the generative system (B loop). Mere
concurrence of both loops separately, however, is not sufficient for
midtended cognition; both loops need to interact, in the sense that
generative interventions by the B loop become nested with the A loop
without the latter losing its intentional character. The agent needs to
\textit{accept} or \textit{adopt} B loop's interventions as part of her
intentional creative process. Then, the whole system becomes a case of
generative \textit{midtended} cognition.

Why and how are AI generated tokens or interventions adopted or
midtended by the agent? At a first approximation, taken in the context
of the examples of word suggestion or text completion, the adoption can
be attributed to the predictive power of the generation. The generated
word(s) are those that the agent was about to use or would have used
given other circumstances (a clearer mind, more time to recall, etc.).
But these circumstances can be amplified further. Generative
interventions might be adopted, or midtended, if the agent finds the
generatively suggested information as a good approximation of what she
could have found herself by doing some research or, and here we enter a
complicated ground, had she read and understood some additional theory
or author (e.g.\ the work of Gilles Deleuze). In fact, generative tokens
can be midtended or made your ``own'', by cognitive proximity, comfortable
familiarity and self-identification, or simply by trust, laziness or
fear that one's own standards are lower than those produced by the AI
system.\footnote{Generative AI systems like Gemini can handle up to 2
million token contexts and be accurately sensitive to their content when
generating the output.}

\subsection{Dimensions of Generative Midtended Cognition}
\label{subsec:dimensions}

\begin{figure}[ht]
  \centering
  \includegraphics[width=0.85\textwidth]{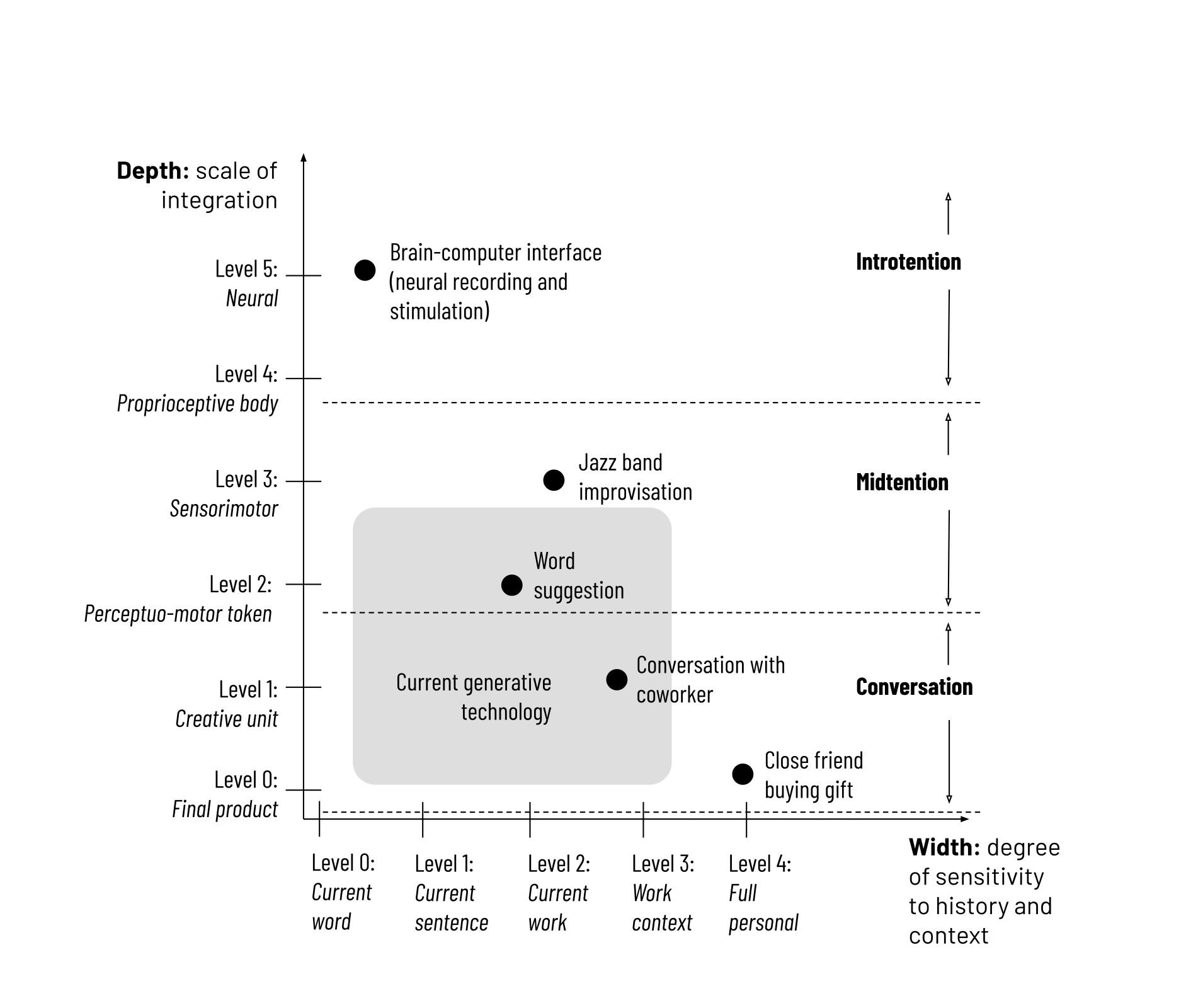}
  \caption{Two main dimensions of generative hybrid space: depth and
  width. The first refers to the degree of reciprocal causal integration.
  The second refers to the length of the context in which Generative AI
  is trained in relation to the agent's history. \textit{[Copyleft 2024
  \textcopyright\ Barandiaran and P\'erez-Verdugo, with CC-by-sa
  licence]}}
  \label{fig:dimensions}
\end{figure}

Inspired by \citeauthor{Heersmink2015}'s (\citeyear{Heersmink2015})
dimensions of integration in extended cognition, we propose that
cognitive processes with intersecting loops of generative power can also
be conceptualised through dimensions of (active)\footnote{We tentatively
term the kind of integration we are targeting ``active'' integration,
instead of the more passive kind of integration theorised in extended
cognition so far. We are aware, however, that this might be misleading
as, in a different context, extended cognition was precisely
characterised as a sort of ``active externalism'' against the ``passive
externalism'' of externalist semantics \citep{Hurley2010}. We have
chosen to maintain it like that for now, given that the ``active''
etiquette there is used in a sufficiently different comparison and
terrain.} integration. Note that whereas Heersmink's (or any traditional
extended cognition literature) account refers to the degree of
integration of external vehicles in a single locus of generative
cognition that is anchored in the agent, we are proposing a different
sort of (active or generative) integration; that of at least two
different loci of generative power---each of which potentially realised
by agent or environmental resources or both---jointly bringing about a
cognitive process. We hereby propose a conceptualization along two
dimensions (see Figure~\ref{fig:dimensions})---although it could
potentially be extended to include more (see footnote~9).

The horizontal dimension, labelled \textit{Width}, involves the degree
of generative sensitivity to the increasingly wider data samples on
which the Generative automaton can be trained, tuned or can take as
context. We assume that the raw foundational systems have been trained
on a massive dataset that includes a huge sample of human generated
digital content (text, video, audio, image, etc.). Beyond that, it can
be increasingly tuned to adapt to the agent's personal history and
context.\footnote{But it might also weaken the stability-inducing effect
that non-AI-reshaped prior intentions had toward future plans, by
increasing resources for practical reasoning and intentional
reconsideration and leaving increasing details of planning open to
artificial generative coordination. In the same vein as social habits
and rituals suffered the destabilizing effect of instantaneous social
communication (e.g.\ submitting social meetings every Wednesday to a
constant negotiation in WhatsApp) we might be at the doorsteps of a
volatile and continuously automatically mediated and re-negotiated
planning under generative AI supported shared intentionality.}

Academic writing can serve as an example to understand what we mean by
this horizontal axis. Level~0 would correspond to the system's
sensitivity covering only the last word typed. Level~1 would include the
whole previous sentence, and level~2 the whole text that is being written
(e.g.\ the whole paper including a first draft of the abstract as sent
to a previous conference, and some notes, quotes and bulleted ideas to
be fully developed). Level~3 would involve contextual access beyond the
current work, including, e.g., the documents you might have selected to
re-read and cite on the paper, the editorial guidelines of the journal,
your teaching notes on the subject, etc. Level~4 includes everything you
have ever written and read (or even from multimodal sources: heard, seen,
etc.).

The second dimension, labelled \textit{Depth}, refers to the scale or
granularity of integration of reciprocal interaction of generative loops.
At level~0 we have a final product (a text, a song, picture, etc.) that
might result from a generative or creative process: e.g.\ a text, a
song, a picture, etc. Level~1 involves conceptual or creative units:
e.g.\ a sentence or paragraph or a musical phrase or rhythm. Level~2
involves perception-action tokens: e.g.\ a word, a note. Level~3
involves sensorimotor dynamics, generally sub-personal or unconscious:
e.g.\ a letter within a text, which we are rarely aware of, neither in
reading nor in writing. If we continue further up within this dimension,
we can get to level~4, the proprioceptive body, where we have to turn to
bodily prostheses as an example. Finally, level~5 would include
integration at the neural scale---potentially achieved by Brain-Computer
Interaction technologies.

This allows us to distinguish three levels of integration along the
depth dimension, one of which targets midtended cognition as we have
defined it. In lower levels along this dimension, where integration
occurs at the level of the final product or creative units, we have
\textit{conversation}, the kind of integration we usually think of when
we think of social interactions. A conversation with colleagues or a
co-authored paper where each author writes a section and then revises
other people's sections would fall within this level. Many of our
current interactions with Generative AI technologies also fall here:
e.g.\ when we give a prompt to an image generator, and seconds later,
without further intervention, it provides an image close to what we
requested, or when we enter into a conversation with ChatGPT. And, as
we can see in the figure, the width dimension nuances the kind of
integration in these conversational cases: there is a clear difference
between a conversation with a coworker and a conversation with a
life-long friend, or between your landlord choosing a new wall paint for
your flat after you asked for renovations and your mum gifting you some
(requested) decoration for your new home.

On the other end, the deepest level, \textit{introtension}, seems to
(luckily) be still quite out of reach, at least for the public. Although
progress has been made in Brain-Computer Interface technologies, and
marketing and hype surround it (see NeuraLink as an example), we have
yet to see integration at this deeper scale to be adopted in somewhat
standard contexts. Our analysis, however, helps forecast its potential
risks and offers grounds for its future theorising.

But we are more interested in the level in between both; the kind of
integration that occurs at scales deeper than complete creative units (an
email, an image, a song), up until the sensorimotor level. The
possibility of this occurring with generative technologies is, we argue,
what remains within the blind spot of previous theorising of extended or
social cognition. As with the conversational level, this does not imply
that midtension only occurs with generative AI. As the participatory
sense-making framework has shown, integration with other agents also
occurs at this deeper scale, and many human intentional processes are,
in a sense, socially midtended. A well documented example of this is the
integration between band members in jazz improvisation (which also
becomes clearly modulated along the width dimension)
\citep{Martinez2017}. The central claim we make is that productive
processes where generative AI is working \textit{within} someone's
creative process, would also fall in a similar category. They are cases
of \textit{generative midtended cognition}.

\section{Should We Stand on the Shoulders of Generative Giants?
Future Risks and Benefits}
\label{sec:future}

Before assessing the future of Generative AI, we need to take into
account that it is a technology that has grown very fast, but that might
be showing signs of fatigue and plateauing. The bending of the
exponential curve is evident for many. It is perfectly possible that,
as a species of technological entity, LLMs and the like might have
reached important limits on their generative capacities. If that is so,
there are different scenarios that can ensue: 1. This is the peak of AI
fantasy, and we are deemed to conform with what we have, or 2. Other
infrastructural or algorithmic innovation might lead to increased
generative capacities. If the first scenario is right, generative
midtension might not expand very far.

The state of the current technology does not afford a comfortable hybrid
generativity. The current problems with hallucinations, reasoning
limitations and deviant chained errors make midtended cognition a
relevant but narrowed phenomenon. In fact, we are aware that in our
analysis we have conceptualised the ``width'' dimension as assuming a
high precision and accuracy. This is definitely an idealisation; the
predictive capacity of different technologies might vary, due to
technical shortcomings, but also to being inaccurate \textit{by design}
(see later discussions about branded content).

However, if the technology keeps evolving, we might increasingly shift
to a cognitive digital culture that relies on generativity as much as
previous civilizations were partially born and sustained by the cognitive
transformations that writing brought about \citep{Ong2012}. Midtended
cognition (and other forms of generative interactions, closer to those
found in functionally equivalent social cognitive tasks) will spread
quickly and become an integral part of our producing and being in the
world.

Technology can evolve in the two dimensions of generativity we outlined
above, towards more granularity on the recursive technological
interventions, and on the increasingly wider contextual tuning or
adaptation. Regarding the first dimension, if cognition emerges out of
recursive loops and, so far, generative technologies only had access to
relatively low resolution external tokens (words, pixels, etc.) the
potential of training, prompting and contextualising the functioning of
generative AI with neuromuscular, proprioceptive, and brain data might
be enormous. The risks of intervening and that scale would be even
bigger.

Perhaps the most promising evolution of the current technology involves
the hybridization of social cognition with generative AI along the
horizontal dimension (including others in the interactive and generative
context). LLMs can facilitate collective intelligence, boosting
midtended agency in the direction of genuine participatory cognition.
But this also poses a delicate risk. Social interaction is to social
autonomy what neuronal interaction is to individual autonomy.
Technologies directed to actively, automatically and massively intervene
in the social interaction fabric, come with high risks. Existing biases
and limitations might well export and be amplified in the social realm,
but the implications go deeper.

Generative AI, when considered from a sociotechnical perspective
\citep[see also][]{Hipolito2023}, moves beyond the cases of individual
creativity we have explored so far; and can become an integral component
of social and political cognition. It is starting to play an increasing
role in mediating and intermediating social interactions. First, by
occupying positions traditionally held by human mediators in social
coordination---e.g.\ as a family therapist mediating in a couple's
conflict, a secretary organizing and scheduling meetings, or a lawyer
mediating between a client and a public institution. Generative AI can
also intermediate between human or machine mediators: e.g.\ automatically
providing meeting schedules for a secretary, bringing up relevant laws
for the lawyer or identifying significant events or \textit{lapses} to
the therapist. More simply, when AI breeds its generative capacity with
you to reply to an email, it is already intervening in social interaction.
This does not only entail a contribution to the information packages we
exchange between humans (emails, lawsuits, confessions or meeting
schedules). By selectively ignoring some aspects of a conversation or
social interaction, by amplifying others, by invigorating politeness or
demanding concreteness, by accelerating response-times or reinforcing
standardized phrases, generative midtension intervenes on the
participatory cognitive interactions that generate and sustain social
norms and regulate social coordination.

Shared intentionality is key to understanding how humans align their
goals and actions to engage in cooperative behaviour \citep{Tomasello2005}.
Generative AI can support these mechanisms by facilitating joint goals
and promoting shared understanding; e.g.\ merging individual and shared
context on midtended generative processes. Moreover, if explicit
intentions serve as guiding frameworks that direct behavior and enable
effective cooperation \citep{Bratman1987}, midtended cognition is called
to play an increasing role in planning, proposing and adjusting
intentions and actions, which allow for smoother alignment of individual
contributions within collective efforts. This generative support for
social adaptability might enhance the fluidity and coherence of social
coordination.\footnote{Sterelny \citeyearpar{Sterelny2010} also focuses
on the ideas of individualization and entrenchment, as opposed to
interchangeability. These concepts highlight the intuition that an
important aspect of integration is the tailoring of the artifact to our
specific cognitive needs so that our cognitive processes become more
efficient. These dimensions have remained relatively undertheorized but
stress the relevance of adapting the environment to the agent's specific
cognitive functions---and, we could say, style---which we will argue
becomes an important area to explore in the case of generative AI.}

Digital infrastructures are an essential part of contemporary political
processes, and they are especially useful for accelerating and scaling
participatory democracy \citep{Barandiaran2024decidim,Berg2021}. AI and
Artificial Life systems can play an important role in empowering these
infrastructures (avoiding the dangers of digital authoritarianism and
centralism) towards models of democratic ecology augmented by (open and
participatory) artificial systems of care and facilitation
\citep{Barandiaran2019}. Generative AI can act as a democratic catalyst,
occupying roles traditionally reserved for human intermediaries---breaking
information bubbles, identifying latent conflicts or facilitating
consensus. In fact, generative AI is already, albeit in an experimental
way, becoming an infrastructural component of social organization,
participating in roles that are integral to democratic deliberative,
propositional and decision-making activities actively synthesizing
collective inputs and (re)writing citizen proposals, that are further
modified and selected by participants, to later bridge the gap between
expert input and participatory engagement, and finally delivering public
policies \citep{Bjarnason2024}.

The risks increase if we consider that, unlike social conquests like
public education and public libraries, the development of generative AI
rests mostly on the private sector and demands increasingly higher
resources that remain out of reach for anyone other than a few bigtech
corporate actors. Moreover, we are still at an early stage of generative
AI service delivery. As the recent history of social media services has
shown, the early stages of corporate success depend on increasing the
user-base and their trust, only to later deploy monetization strategies
that are mostly directed at commoditizing human behaviour \citep{Zuboff2019}.
The commodification of human intentionality through its generative
midtended extension might be one of the deepest risks we face in the
near future of AI's sociotechnical expansion. If the \textit{attention
economy} is already showing its devastating effect on human autonomy and
well-being \citep{Bhargava2020,GonzalezDelaTorre2024,Williams2018}, it
is hard to imagine the potential negative impact of an
\textit{intention economy} deployed through generative human cognitive
steering. The attention economy has so far mostly developed by grabbing
our attention and influencing future behaviour by shaping or creating
desires (e.g.\ TV advertisement) or by conditioning of facilitating
attention-action possibilities (online clicking to buy a product). If
human intentionality however is increasingly meshed with midtended
generative processes as we have defined it, the way is open for
commodifying the generative process.

Going back to the opening example of word suggestion, we can find a
parallel example that illustrates the intention economy to come. When
the waiter comes, and you are about to complete your order, you hesitate
with the wine. ``We are having fish, and it is a hot day, how about a
fresh white wine, like\ldots'' when suddenly you hear ``Txakoli
Generative''. It is not a suggestion of your partner nor the waiter's.
It is the voice that comes out of the speaker of an AI embedding glasses
(that include cameras that capture everything you see and a small
speaker to your ears). The AI, however, knows both your tastes and the
current context, i.e.\ the meal you are about to order, as you
discussed it with your meal colleagues, and images of the long wine cart
that overwhelmed you. Whispering the right choice, at the right time, in
the right context, is a way to commodify midtension. This opens the way
for an economy of intention. Like the word suggested by a friend and
integrated into your discourse about Deleuze, products, places or
services might also be recurrently injected into your verbal or
interactive intention generating spaces. Is Txakoli Generative the wine
you ``really wanted''? Well, you didn't fully know, or maybe you did,
you tried it in the past and, in fact, it passed unnoticed to you on the
wine chart. Had you known it was there, you might have chosen it
yourself. But the same might have been the case of the other 2 wines on
the list. Yet, Txakoli Generative paid a premium plan for its promotion
on the intention economy market for the AI embedded glasses you are
wearing. More examples can be added to illustrate this tendency.
Following the manifestation of generative midtension in writing an
academic paper, it is not difficult to think of the generative assistant
suggesting references from authors of universities that have paid a
marketing-package on their subscription.

Beyond this, even if we dodged those risks, one further line of worry
that has already been brought up in the literature
\citep{HernandezOrallo2019} is that of a sort of \textit{cognitive
atrophy}. Motor vehicles induce movement atrophies that we need to
compensate with gyms and sport habits. Calculators severely reduced our
capacity to carry out fast and efficient mathematical calculations. What
kinds of creative atrophies might result from generative midtended
cognition in humans? We also need to take into account that not only
humans are exposed to creative damage. Generative AI can also collapse
when recursively trained and tuned to cannibalised content
\citep{Shumailov2024}. Could a generalized adoption of generative AI
result in a global creative atrophy, incapable of generating sufficiently
rich and original content to feed itself forward?

Finally, issues of authenticity \citep{Zawadzki2021}, which are already
at stake within current digital contexts \citep[see][for an exploration
within an enactive framework]{PerezVerdugo2023}, might also get amplified
with generative midtension. If personal identity is the result of our
creative and narrative capacity to write and tell ourselves in the world
\citep{Ricoeur1995}, and generative interventions occur at scales that
usually fall outside our awareness by technologies that we are unable to
regulate (since they are mostly opaque for us), our personal autonomy
and identity will be at stake. While the user can retain a sense of
authorship over the final product, as she is the only true agent,
midtended cognition implies that this does not guarantee that it is a
completely authentic product of her agency in the traditional sense. To
assess the extent of authenticity, the width dimension could be called
in to shed light on some nuances: even when integration is deep, we
cannot ascribe the same degree of authenticity to a wider midtended
process where generative AI is tuned to the level of full personal
context of the user, than to shallower cases where the tuning is more
local. This is also the intuition that the individualization dimension
of integration in extended cognition was tapping upon.

With generative AI, however, this can be misleading. While in the
dimensions of individualization of extended cognition the tuning is
assumed to be made and regulated by the human agent, here the tuning
relies on the generative AI itself---or rather the companies that develop
them. This means that it can be more difficult to elucidate the true
scope of the tuning, beyond our definition of width: a particular LLM
could be tuned to my full personal context \textit{and} to the context
of the company that developed it, that has a particular set of values
and goals different from mine---something I cannot be aware of (unless
the system is fully open-source, explainable and properly audited by
trustworthy agents). I might find myself using a seemingly
hyper-integrated AI technology (think of a future personal generative AI
that accompanies you during your life to be continually trained in the
whole history of your creative outputs and processes, and that is
integrated at the deep level of perceptuo-motor token generation) that
nevertheless is also inadvertently trained on branded content for an
investor of the developer company. The authenticity of my outputs will
surely be at stake. Future developments of the work we propose here,
such as finer analytical work on the width dimension or the
incorporation of further dimensions, are necessary to better account for
current and future cases of generative AI use.

Will we stand on the shoulders of generative AI? If we understand, as
\citeauthor{BarandiAran2024arxiv} suggest, that LLMs operate as a
library-that-talks (or a media-library that generates ``new'' media),
generative midtended cognition implies an automatic standing on the
writings of others. And this is already becoming an important productive
and creative boost for those that have the opportunity to explore it. In
this sense, the most optimistic conceptions of the future of generative
AI consider its democratising potential, given its promise of levelling
anybody up to an interdisciplinary PhD or expert level intelligence. But
we might also soon witness the emergence of a ``generative class'' whose
main goal is to co-pilot generative machines within an assembly network
of digital intelligence processes. Deep forms of alienation might ensue
also at the productive side of generative cognition (not only at the
``consumerism'' and marketing-targeted side), analogous to, yet
psychically deeper than, those that industrial capitalism brought with
it.

However, this can also be turned around. An unexpected (certainly not
industry-promoted) form of benefit from generative AI could be that of
``authentic creativity through negative dialectics with generative AI''.
It is possible to let yourself go by AI generative waves, but, on the
contrary, it might be equally possible to surf or ride the waves tricking
against the flow by a sort of negative dialectics: to take impulse on
the generative wave just to twist direction and find creativity always
\textit{against} what the AI has generated (as suggested by
\citealt{Rushkoff2024}). If instead of a conservative repetition of what
was done in the past (by oneself and others), the ultimate and deepest
sign of authenticity is to be genuinely open to change and becoming,
then generative AI might boost the means to become more openly authentic
by dialectically building against generically and even
personally-tuned generative suggestions.

\section{Conclusion: Active Integration Beyond Extension}
\label{sec:conclusion}

The framework of generative midtended cognition that we have proposed
aims to provide a theoretical basis to capture the hybrid processes
where AI-generated suggestions become integral to the intentional
creation of cognitive products by human agents. Understanding the ways
in which artificial generative technologies can become integrated in our
cognitive processes, without admitting to them being full-fledged cases
of social interaction or of extended cognition as theorised so far,
becomes a crucial step to analysing our relationship with AI. By
articulating the entanglement of different nested loops of generative
power that contribute to a shared cognitive creative outcome, we have
conceptualised two dimensions of active integration, width and depth.
In doing so we have been able to characterise the specific case of
generative cognitive processes occurring at a scale that was not
displayed in our relations with previous forms of technologies: that of
\textit{midtended} cognition.

The novelty of our approach lies in the fact that previously existing
theories of extended (or enactive) cognition did not foresee that the
environment would be populated by the generative technologies we now
have at our disposal. The environment was not thought of as being
\textit{generative} in the relevant sense (a purpose-structured kind of
generativity similar to human creative, purposeful, practices) nor tuned
to the specific context and agent in its generativity. This is not only
the case for extended cognition theorising; postphenomenology,
particularly as developed by Verbeek, has advanced a strong paradigm on
understanding hybrid (human-technology) kinds of intentionality.
\citeauthor{Verbeek2008}'s (\citeyear{Verbeek2008}) account of cyborg
and composite intentionality aims at something similar to our account
of integration, since he recognises the intentional character of
technology. However, the sort of intentionality instantiated by
technology, in his account, is always directed towards the world, and
not back to the agent. This account falls short of capturing current
generative technology, where the arrow of (derived) ``intentionality''
(or purpose-structured generativity, in our account) departing from
technology would also point towards the human agent.

Analysing the challenges and implications of these technologies through
the lenses of these previous approaches (as developed so far) would tend
to attribute all purpose-structured or normatively shaped contributions
to the human agent, with the AI tool functioning only as a vehicle or a
constraint in its realization, or would fail to capture the uniquely
generative character of the products of the interaction. This misses the
importance of the contributions of the AI technology in the generative
process. As such, our approach allows us to further analyse the ethical
risks that come with digital technologies and that were already starting
to be raised within extended cognition literature. Coming back to the
issue of transparency, for instance, our dimension of depth allows us to
account for cases where the scale of interventions occurring below the
creative unit grounds the phenomenological feel of transparency, while
still retaining the fact that the cognitive process is a result of two
different sources of generative power. The creative units that emerge
from these hybrid processes (for instance, a sentence created with an
autocomplete function) are a result of midtended cognition, not merely
extended cognition.

On the other hand, treating these processes as true forms of social
cognition would obscure the fact that generative AI technologies, while
capable of producing generative outcomes, are not autonomous agents
capable of participating in a true form of social interaction, with its
constitutive tensions \citep[see][]{DiPaolo2018}. As such, our capacity
to negotiate our interactions with these kinds of automata are not
similar to what occurs with other human agents. While this does not
devoid midtended cognitive processes that include AI technologies of
their creative or generative character, it does have important
implications for how we might be able to (collectively and individually)
regulate them.

We have aimed, in the previous section, to hint at the specific risks,
questions and possibilities that generative AI can pose if we understand
them as bringing about midtended cognition processes. The granularity of
the intervention in the joint generative process (captured by our depth
dimension), together with the immense capacity for context-sensitivity
(width dimension) and the status of the genAI automata as (currently
market-driven) artificial products, has brought about specific concerns
that are unprecedented in the history of technology. The political
strategies needed to navigate these issues are different from those used
to overcome issues emerging in purely social interactions between
autonomous agents (but also from passive, non-generative technologies,
as those amenable to extended-mind style theorizing). And we have shown
that we do not need to consider LLMs as possessing human qualities to
account for the fact that generative AI can intervene on cognitive
processes in a generative manner not only at the conversational level,
but also at deeper levels in the shaping of the process.

We need new concepts to talk about our relation with and use of
generative AI. Cognitive science, and extended cognition in particular
as the vantage tradition in including the environment in analyses and
descriptions of cognitive processes, has a strong (political,
philosophical and scientific) responsibility in offering analytical
tools to assess not only the mode of existence of generative AI
\citep{BarandiAran2024arxiv}, but also our modes of interactions with
it; its risks and its benefits.

\bibliographystyle{plainnat}
\bibliography{references}

\end{document}